\documentclass[11pt]{article}

\usepackage{preamble}
\usepackage{multicol}
\usepackage{color}

\title{Symmetry Group Equivariant Architectures for Physics \\ \Large A Snowmass 2022 White Paper}

\author[1]{Alexander Bogatskiy}
\author[2]{Sanmay Ganguly}
\author[3]{Thomas Kipf}
\author[4]{Risi Kondor}
\author[4]{David W. Miller\thanks{Contact Editor, David.W.Miller@uchicago.edu}}
\author[5]{Daniel Murnane}
\author[4]{Jan T. Offermann}
\author[5]{Mariel Pettee\thanks{Contact Editor, mpettee@lbl.gov}}
\author[6]{Phiala Shanahan}
\author[7]{Chase Shimmin}
\author[8]{Savannah Thais}

\affil[1]{Flatiron Institute}
\affil[2]{ICEPP, University of Tokyo}
\affil[3]{Google Research}
\affil[4]{University of Chicago}
\affil[5]{Lawrence Berkeley National Laboratory}
\affil[6]{Massachusetts Institute of Technology}
\affil[7]{Yale University} 
\affil[8]{Princeton University}

\usepackage[firstpage=true]{background}
\backgroundsetup{contents={\parbox{6.5in}{\snowmass}}, scale=1,placement=top,opacity=1,color=black,position={3.25in,1.2in}}

\begin{document}

\maketitle

\begin{abstract}
Physical theories grounded in mathematical symmetries are an essential component of our understanding of a wide range of properties of the universe. Similarly, in the domain of machine learning, an awareness of symmetries such as rotation or permutation invariance has driven impressive performance breakthroughs in computer vision, natural language processing, and other important applications. In this report, we argue that both the physics community and the broader machine learning community have much to understand and potentially to gain from a deeper investment in research concerning symmetry group equivariant machine learning architectures. For some applications, the introduction of symmetries into the fundamental structural design can yield models that are more economical (i.e. contain fewer, but more expressive, learned parameters), interpretable (i.e. more explainable or directly mappable to physical quantities), and/or trainable (i.e. more efficient in both data and computational requirements). We discuss various figures of merit for evaluating these models as well as some potential benefits and limitations of these methods for a variety of physics applications. Research and investment into these approaches will lay the foundation for future architectures that are potentially more robust under new computational paradigms and will provide a richer description of the physical systems to which they are applied.
\end{abstract}

\clearpage
%\snowmass

{
    \hypersetup{linkcolor=black}
    \tableofcontents
}

%=============================
\section{Executive Summary}
\label{sec:executive}
%=============================

% \editnote[DWM]{We will put a very short executive summary here in order to aid the Topical Group Conveners and the Frontier Conveners in writing their summary reports.}

This White Paper highlights the potentially significant scientific and practical benefits of developing machine learning architectures for current and future physics applications that are inherently organized around symmetry considerations and structures. %This Executive Summary briefly describes the context of this work as well as the challenges, opportunities, and potential benefits that pursuing these concepts may have for physics.

%Symmetries underlie countless physical phenomena throughout the universe across all length and energy scales. Mathematical descriptions of the symmetries that we observe in nature have been invaluable in inspiring some of the most successful physics theories to date. Incorporating knowledge of these symmetries and their equivariance properties into the algorithms and architectures that we use to investigate and describe physical systems appears to have numerous advantages and yet also introduces constraints and potential limitations. Rotation, translation, and temporal symmetries, for example, have already been explicitly included in machine learning architectures have resulted in significant performance benefits for models such as convolutional neural networks (CNNs), recurrent neural networks (RNNs), graph neural networks (GNNs), and transformers. However, despite the reliance of theoretical physics as a discipline on a variety of symmetry groups, the explicit inclusion of these symmetries into machine learning models designed for physics datasets is rare. 

%-----------------------------
\begin{itemize}
    \item Symmetries underlie countless physical phenomena throughout the universe across all length and energy scales. Mathematical descriptions of these symmetries that we observe in nature have been invaluable in inspiring some of the most successful physics theories to date.
    \item Incorporating some kinds of symmetries (rotational, translational, sequential, temporal, etc.) into machine learning architectures has already resulted in significant performance benefits for models such as convolutional neural networks (CNNs), recurrent neural networks (RNNs), graph neural networks (GNNs), and  transformers.
    \item Despite the reliance of theoretical physics as a discipline on a variety of symmetry groups, the explicit inclusion of these symmetries into models designed for physics datasets is rare.  
    \item The potential benefits to physics applications with respect to non-equivariant models range from \textbf{model performance} (e.g. better generalizability) to \textbf{resource efficiency during training} (e.g. achieving strong performance with a highly reduced training dataset and/or model parameters) to \textbf{interpretability} (e.g. potentially mappable to meaningful physical observables). %We can even realize some of these benefits for use cases with approximate symmetries. 
    \item The investment of time and funding into the construction of equivariant architectures for physically-motivated symmetry groups will have far-reaching benefits for applications across fundamental research in science and industry. \textbf{The development of symmetry group equivariant machine learning architectures has substantial potential to aid in future physics discoveries, but only if we recognize its importance by dedicating the necessary resources to support it.}
\end{itemize}
%-----------------------------

%=============================
\section{Introduction \label{sec:intro}}
%=============================

% \editnote[Editors]{These are the major bullet points that we intend to address with the introduction.    
%     \begin{itemize}
%         \item Description of equivariance and symmetry groups broadly
%         \item Examples of implementations in machine learning architectures
%         \item Description of the potential benefits and drawbacks
%         \item Statement of the open problem(s) in exploring this space
%         \item Call to action
%     \end{itemize}
%     Still need to make the intro more concise and ensure we are making these points.
% }    
    
    Nearly all areas of science and industry are witnessing a renaissance in the development, application, and impacts of machine learning (ML) applied to both old and new problems. Sophisticated and extremely diverse sets of algorithms based on a wide array of ML architectures and design principles are breathing new life into problems ranging from image recognition~\cite{10.1162/neco_a_00990} to silicon sensor readout~\cite{Aad:2014yva}, from intelligent molecular design~\cite{SCHNEIDER1998175} to parsing the landscape of string theories~\cite{Carifio:2017bov} and accelerating theory calculations~\cite{Albergo:2019eim}, and from distinguishing quarks from gluons~\cite{Komiske:2016rsd} to controlling telescopes searching the cosmos~\cite{sandler_use_1991}. As the computational and architectural landscape of ML continues to evolve, it is essential to also reassess the fundamental mathematical structures that underlie these approaches, especially when they are applied to the modeling of physical systems. 

    %The introduction of symmetries into the fundamental mathematical structure of these architectures has the potential to .... -- in the case of convolutional neural networks (CNN), translation symmetry or invariance. 
    
    One of the driving principles that has often guided the emergence of new directions in theoretical physics and placed constraints on the methods used to test them is \textit{symmetry}. Global and gauge symmetries, continuous and discrete symmetries, internal and external symmetries: the existence of any one of these can place constraints on the dynamics of a system, require conservation of quantities and quantum numbers, or reduce the dimensionality of the phase space in which the system evolves. However, it is rare to see the explicit inclusion of these symmetries into the ML algorithms used to model the dynamics of particles or the evolution of the universe, for example, despite the fact that symmetries are an inherent and often defining characteristic of these systems and processes.
    
    In this White Paper we highlight the potentially significant scientific and practical benefits of developing machine learning architectures for current and future physics applications that are inherently organized around symmetry considerations and structures. Building these customized architectures for physics use cases will require special prioritization and investment from our community, as this pioneering work often falls outside of the scope of traditional research grants. Given the strong motivations supporting the potential of equivariant models to contribute to our scientific discoveries, we cannot afford to miss out on the opportunity to deepen our understanding of the relationship between physics and machine learning through the lens of fundamental symmetries. 
    
%=============================
\section{Equivariance in Machine Learning \label{sec:equivariance}}
%=============================    
    
    %% Removed mention of the technical term ``group'' until the next paragraph ~CS
    Most problems in physics involve structured data that have some inherent compatibility with symmetries: Euclidean vectors, Minkowski vectors, indistinguishable particles, and so on.  Similarly, the underlying processes generating the data of interest, such as particle collisions or a Monte Carlo simulation, possess definite symmetry properties.  These problems may require or at least benefit from models that construct intermediate representations and perform computations that reflect the symmetry in question.  Following this approach, elegant architectures can be informed by these principles, and the ``building blocks'' of such architectures may be limited to those allowed by the imposed symmetries.  Beyond imbuing the network design with properties that mirror the physical system that it describes, this may -- perhaps counter-intuitively -- improve out-of-distribution generalization, interpretability, and uncertainty quantification, while also allowing for certain simplifications of the model itself. Carefully and appropriately implemented, this may therefore be a highly sought-after property in neural network design.
    
\begin{figure}
        \centering
        \includegraphics[width=\textwidth]{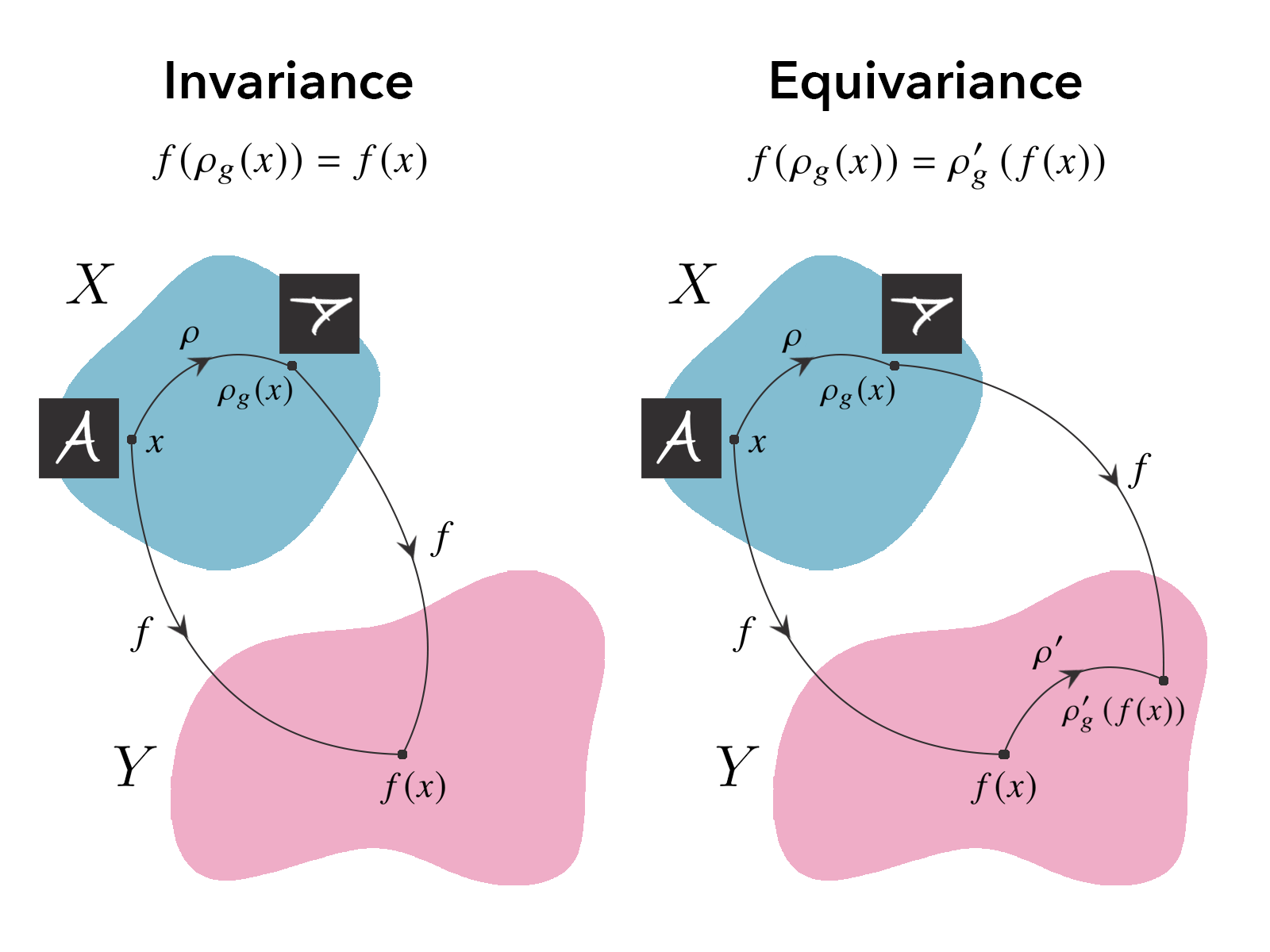}
        \caption{An illustration of the differences between symmetry group invariance and equivariance for the example case of identifying a handwritten letter in an image. Here, $f: X \rightarrow Y$ is a map between vector spaces $X$ and $Y$. $\rho_g(x) \equiv \rho(g,x)$ is an action of a group $G$ on $X$ and $\rho'_g(y) \equiv\rho'(g,y)$ is an action of a group $G$ on $Y$. The \emph{invariant} model (left) will output the same result on both the original and translated images, while the \emph{equivariant} model (right) will transform the translated image in a way that reflects the underlying symmetry group. More formally, this means that the map $f$ is equivariant with respect to the actions $\rho:G\times X\to X$ and $\rho':G\times Y\to Y$ if $f(\rho_g(x)) = \rho_g' \left(f(x)\right)$ for all $x \in X$ and $g \in G$.}
        \label{fig:inv_equiv}
    \end{figure}

    %% I think this the first mention of equivariance, it should be defined fairly early on ~CS
    Mathematically, symmetries are usually described by \textit{groups}.
    We can characterize the relationship between a function (such as a neural network layer) and a symmetry group by considering its \textit{equivariance} properties.
    A map $f: X \rightarrow Y$ is said to be equivariant \textit{w.r.t.} the actions $\rho:G\times X\to X$ and $\rho':G\times Y\to Y$ of a group $G$ on $X$ and $Y$ if
    %-----------------------------
    \begin{equation}\label{eq: equi}
        f(\rho_g(x)) = \rho_g' \left(f(x)\right)\,
    \end{equation}
    %-----------------------------

    \noindent for all $x \in X$ and $g \in G$. Here, the notation is $\rho_g(x)\equiv \rho(g,x)$. Most commonly we are concerned with the case where $X$ and $Y$ are vector spaces and 
    $\rho_g$ and $\rho'_g$ are linear maps, in which case $\{\rho_g\}$ and $\{\rho'_g\}$ are so-called   \emph{linear representations} of $G$.  
%    are expected to return matrices $\rho_g \in \mathrm{GL}(X), \rho_g'\in \mathrm{GL}(Y)$.
    Intuitively, Equation \ref{eq: equi} tells us that for a given transformation of the input, the output of an equivariant function transforms in a definite way that preserves the group structure.\footnote{This is inherent in the definition of the group representation: $\rho_g \rho_h \cdot f(x) = \rho_{gh}' \cdot f(x)$.}
    Note that in the special case where $\rho_g$ is the identity for all $g\in G$, $f$ is said to be \textit{invariant} with respect to $G$. A diagram illustrating the distinction between invariance and equivariance can be found in Figure \ref{fig:inv_equiv}. 
    
    %-----------------------------
    %-----------------------------
    
    %% Concrete example using the familiar CNN, might be a bit long... ~CS
    At first, it may seem that invariance would be the most interesting form of equivariance.
    For example, consider the problem of recognizing a letter at an arbitrary location within an image.
    The data contains some continuous variation due to a translation symmetry, yet it is desirable that an image recognition algorithm meant to identify that letter -- also referred to as a \textit{classifier} -- be able to do so accurately and regardless of the letter's position.
    In this case, the classification algorithm should be invariant with respect the group of translations.
    However, the overwhelming success of Convolutional Neural Networks (CNNs)~\cite{NIPS1989_53c3bce6, 10.1162/neco_a_00990} in image processing serves to illustrate that even when an invariant is desired for the final output, it is beneficial to construct more general equivariant intermediate representations.
    CNN layers are not invariant, but rather preserve discrete translation symmetries at each layer; if the input image is shifted by a certain number of pixels, the output of the CNN layer (ignoring boundary effects) is shifted by a corresponding amount~\cite{CohenWelli16, dieleman2016exploiting}. In a typical CNN, an invariant representation is typically formed only in the last few layers of the network.

    Intuitively, the reason CNNs are so effective on image data is because they are able to learn specific features, such as edges or textures, that can be matched at any location on the input image.
    This economizes parameters, since the CNN need not learn the same features over and over at every location.
    This also improves generalization, since a given feature can be detected even if it appears at a location that was not observed in the training dataset.
    
    %% Segue to physics-specific motivation of equivariance ~CS
    This motivates us to consider whether new architectures can have a similar impact by leveraging symmetries relevant to problems in physics, just as CNNs did by exploiting the natural translation symmetry of image data, and doing so by constructing equivariant latent representations of the inputs at every layer of the network. In the context of problems with exact symmetries, it therefore makes sense to consider architectures that are fully made up from \textit{only} equivariant operations.
    To this end, we must design ``building blocks'' with equivariance properties corresponding to the symmetries of interest. Note that various deviations from this approach cannot be dismissed, as will be discussed in Section \ref{sec:softequivariance}.
    
    The overall viability of equivariant machine learning can be motivated by mathematical results such as universality theorems. The most general form of equivariant linear layers with respect to the action of any compact group was derived in \cite{KondoTrive18}. In \cite{Yarotsky18} it was shown that for a wide class of Lie Groups, all equivariant mappings can be expressed in terms of a finite basis of invariants and equivariant tensors and easily approximated with conventional feed-forward neural networks. Further, in \cite{Kondor2018Nbody, BoAnOfRoMK20} it was shown that a fairly generic equivariant architecture based on tensor products is sufficient to implicitly generate the relevant finite basis. Thus, equivariance does not impose any real constraints on the overall design of neural networks, other than requiring all node-level operations to be equivariant. In particular, recurrent, adversarial, generative, etc. architectures can be constructed out of the equivariant ``building blocks'', easily mimicking traditional architectures.
    
    These general ideas have already led to the development of multiple equivariant architectures for sets (permutation invariance)~\cite{ZaKoRaPoSS17, dolan}, graphs (graph isomorphisms and permutation equivariance)~\cite{maron2018invariant}, 3D data (spatial rotations)~\cite{Monti2017Geometric,thomas2018tensor}, homogeneous spaces of Lie groups such as the two-dimensional sphere~\cite{CohenSphericalICLR2018,KondLinTri18}, and even gauge equivariant systems~\cite{Kanwar:2020xzo,Boyda:2020hsi}. In fact, the success of many of the more traditional architectures like CNNs and Recurrent Neural Networks (RNN) can be largely attributed to their equivariance properties~\cite{10.1162/neco_a_00990}.
    
    Symmetries play a central role in any area of physics~\cite{Frankel04}, and as such physics may provide one of the widest variety of symmetry groups relevant in computational problems. In particular, high-energy particle physics involves symmetry groups ranging from $\mathrm{U}(1)$, $\SU(2)$ and $\SU(3)$ to the orthochronous Lorentz group $\SO^{+}(1,3)$, and even more exotic ones like $\mathrm{E}_8$ and conformal Lie algebras. Architectures that respect these symmetries may be able to provide more sensible and tractable models with parameters that could be directly interpreted in the context of known physical models, similar to attention mapping studies for CNNs used in image recognition. 
    
    %Harmonic analysis provides two parallel but closely related implementations of group equivariance in neural networks. The first is a natural generalization of CNNs to arbitrary Lie groups and their homogeneous spaces \cite{CohenWelli16}, where activations are functions on the group, the nonlinearity is applied point-wise, and the convolution is an integral over the Lie group. The second approach works entirely in the Fourier space \cite{ThSmKeYLKR18,AndeHyKon19}, that is, on the set of irreducible linear representations of the group, and the nonlinearity is based on tensor products and Clebsch-Gordan decompositions.
    
    These ideas are general and may be applied in any field of computational physics. In this White Paper we focus on a particular subset of applications in particle physics where the data typically contain the energy-momentum 4-vectors of particles produced in collision events at high-energy particle accelerators, or by simulation software used to model the collision events. As 4-vectors, these data naturally support an action of the Lorentz group, therefore architectures that are Lorentz-invariant or Lorentz-equivariant are a major step towards understanding the ways in which machine learning can solve physical problems with such inputs. The first successful application was for the Lorentz-invariant task of top-tagging~\cite{BoAnOfRoMK20}, where a competitive performance was reached despite a dramatic decrease in the number of parameters. We are confident that in the near future these applications will be extended to complex regression tasks and tasks with vector targets.

%=============================
\section{Potential Benefits of Equivariant Models \label{sec:benefits}}
%=============================
% \editnote[Editors]{These are the major bullet points that we intend to address with this Section.        
% \begin{itemize}
%     \item Simplicity: reduce overall number of parameters
%     \item Data efficiency
%     \item Inductive biases
%     \item Interpretability 
%     \item Uncertainty: impact on training, final results, etc.  
%     \item \textbf{KNOWN} benefits: AlphaFold, computational chemistry/protein folding, etc.
% \end{itemize}
% }
    
    The success of neural networks is often attributed to the massive computational capacities of modern computers, allowing one to practically optimize models with billions of parameters. Mathematical statements like the Universal Approximation Theorem~\cite{HORNIK1989359}, while not guaranteeing much in practice, make it plausible that an arbitrarily complex system can be modeled and learned even by a simple fully-connected neural network. Nevertheless, it is likely that such ``uninformed'' architectures which make no reference to the underlying structure of the system will not yield efficient nor possibly even optimal models even if they can, in principle, provide a high accuracy of predictions. This section elaborates on the potential benefits to consider in the construction of more ``informed'' architectures. We also discuss why a focus merely on the accuracy of predictions can be detrimental to machine learning in scientific contexts, and how it can be re-balanced for more physics-oriented model development.
    
    As discussed in \Secref{equivariance}, many architectures are successful precisely because they take advantage of some intrinsic properties of the corresponding system under study -- for instance, spatial translation invariance of image recognition, `time' translation equivariance in text, and permutation equivariance in graph networks. Recently, generalizations of the symmetry group of the convolution operation~\cite{finzi2020generalizing} as well as message passing graph neural networks~\cite{satorras2022en} show that the complexity of the learning task can be significantly improved with fewer parameters. Symmetry equivariant transformer networks can surpass the performance of the corresponding convolutional networks~\cite{hutchinson2021lietransformer}, and symmetry equivariant normalizing flows can produce better generated outputs than non-equivariant normalizing flows~\cite{satorras2105.09016}. It has been demonstrated that, in general, the symmetry equivariant networks can estimate conserved quantities associated with the dynamics better, compared to a general fully connected network~\cite{finzi2020generalizing,batzner2021e3equivariant}. Some use cases such as accelerating first-principles theoretical physics calculations suggest that building in symmetries might be required for guarantees of exactness, or to properly encode the theory. For the application of generative models to accelerate lattice gauge field theory calculations, incorporating the high-dimensional gauge symmetry exactly was found to be essential to train models~\cite{Kanwar:2020xzo,Boyda:2020hsi}. We may conclude that the value of such architectures lies not just in the quality of the predictions they produce, but also in a number of other properties that make them better \textit{models}, as opposed to simply better \textit{predictors}:
    
    %-----------------------------
    \begin{itemize}
        \item \textit{Model size/complexity.} With the symmetries of the underlying problem built-in, the model size is constrained with respect to one that does not respect these symmetries. For a given number of parameters, the symmetric model may use fewer parameters more efficiently~\cite{CohenWelli17}. Lower-dimensional models may also be desirable if they are to be used as physical models, but model size is also crucial in many low-latency applications. Fewer parameters may also result in a less complex loss landscape and thus more efficient optimization and training dynamics. \Secref{softequivariance} expands on this topic with a discussion of approximate and partial symmetries.  %\editnote[TK]{This final statement might be controversial, since there appears to be anecdotal evidence that larger-scale models are not necessarily more difficult/unstable to train, often we can indeed observe the opposite: having more parameters appears to allow a model to escape a bad local minimum more effectively. See also the paper on "Why do Larger Models Generalize Better?" for a related discussion.}
        \item \textit{Interpretability.} A model constrained by the symmetry group(s) of the underlying problem may be more interpretable than a general one -- not only for likely having fewer parameters, but also because these parameters will represent physically meaningful \textit{observables}. In the context of CNNs, this fact is used when interpreting convolutional kernels as translation-invariant visual features. In a more general context, the parameters of a symmetric model can be made themselves invariant, and therefore potentially direct representations of measurable quantities. Additionally, by comparing equivariant and non-equivariant architectures applied to the same problem, one may be able to more easily probe the learning dynamics of the non-equivariant model and characterize what non-symmetric information is being learned.
        \item \textit{Sample efficiency.} In contrast to the common way of training a generic unconstrained model to respect symmetries, i.e., by \textit{augmenting} the dataset with a multitude of symmetry transformations, an equivariant model can potentially achieve the same performance with a dramatically reduced training dataset. For example, training a fully-connected network for image recognition may require many modified copies of the input data for the network to ``learn" a translational or rotational symmetry, whereas a CNN requires fewer such samples. Improvements in sample efficiency have been shown to range from factors of a few to thousands~\cite{batzner2021e3equivariant,gong2022efficient,dax2021group}.
        \item \textit{Generalizability.} If an equivariant model can learn an entire \textit{orbit} (manifold spanned by the action of the symmetry group) from a single sample, then, similarly, any new input from the same orbit will produce an output entirely determined by equivariance. In essence, symmetry allows the network to learn exact reduced representations, and dimension reduction is a known method of enhancing generalization []. Additionally, diverse training paradigms can be applied to a single equivariant model architecture in order to extend the generalizability of the model depending on a variety of end use cases. For instance, if training efficiency is a priority, one could include symmetry-breaking parameters initially during the training that subsequently decay, thereby allowing a more efficient optimization pathway through nonphysical parts of model space. Alternatively, for systems with approximate symmetries, one could start with an exact equivariance requirement that is then loosened later in training.
        \item \textit{Faithfulness to physical laws.} Finally, since equivariant architectures  implement physical symmetries as hard constraints ``baked into'' the structure of the network itself, they can guarantee that whatever is learned by the model is not going to violate known laws of nature. This is particularly important in, e.g., quantum chemistry applications, where an incorrectly learned interaction term could for example violate the law of conservation of angular momentum.
    \end{itemize}
    %-----------------------------
    
While equivariant models are not yet commonplace in fundamental physics and astrophysics, other fields such as computational chemistry, protein folding, and more have seen clear benefits from these architectures -- with one recent and notable example being the success of AlphaFold~\cite{alphafold} in protein structure predictions.

\begin{figure}
    \centering
    \includegraphics[width=0.7\textwidth]{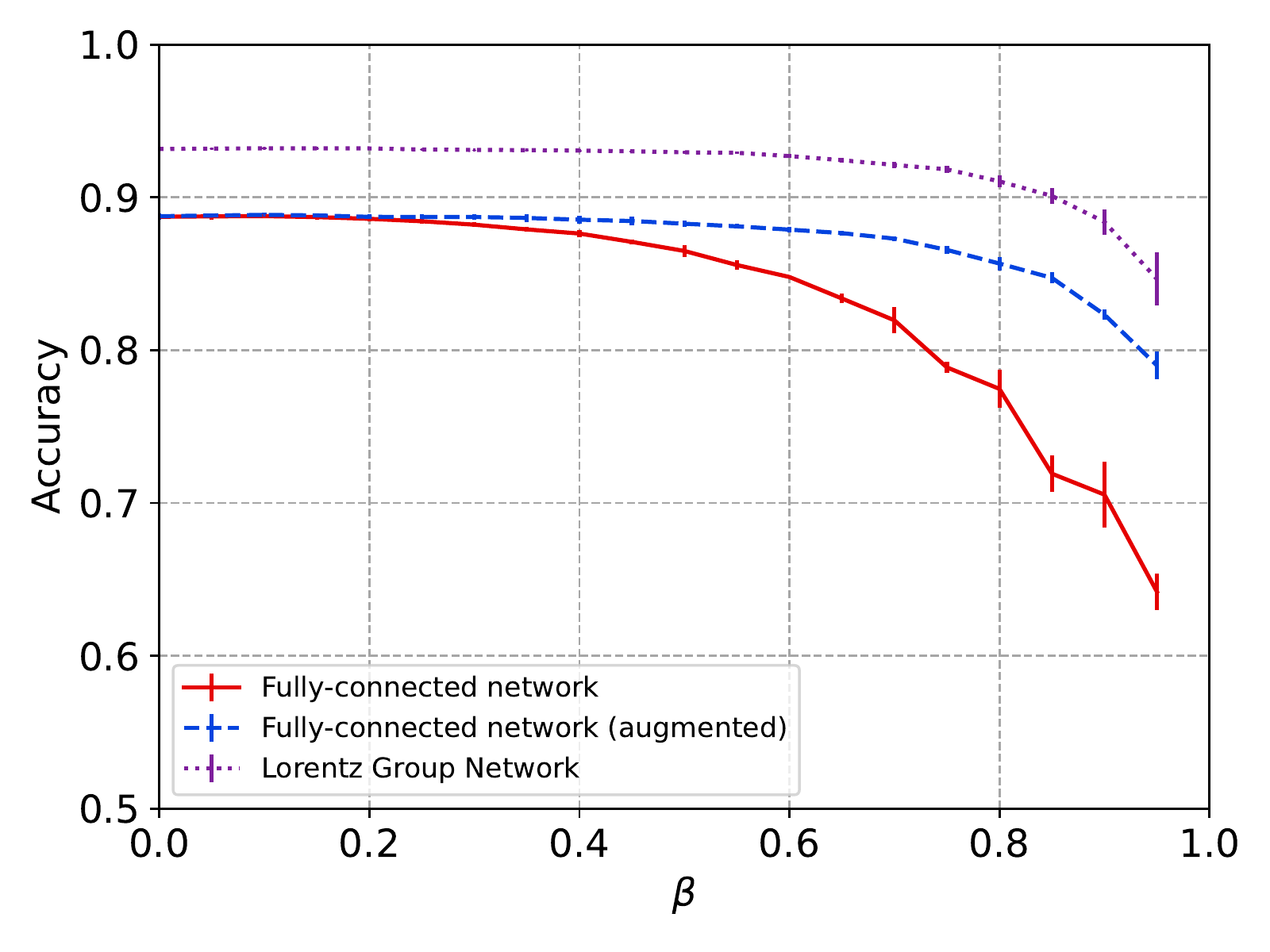}
    \caption{The accuracy of various neural networks in performing top quark tagging, as a function of a Lorentz boost applied to the testing data reference frame (parametrized with $\beta = v/c$). The simple, 4-layer fully-connected network is not robust against Lorentz boosts, so the classification accuracy suffers as the testing data has a hidden Lorentz boost applied. While this can be potentially mitigated by careful training data augmentation -- in this case by applying random Lorentz boosts to training events -- an equivariant architecture like the Lorentz Group Network~\cite{BoAnOfRoMK20} can provide for robustness without the need for any special training procedure. Decreased accuracy at relatively high boosts for LGN is due to numerical precision limits of that particular design. Network training and testing was performed using the top-tagging reference dataset~\cite{KasPleThRu19}, and each curve is averaged over three separately trained network instances.}
    \label{fig:boost_invariance}
\end{figure}

Studying the above qualities of a model is likely to benefit both ML practitioners and physics phenomenologists. For example, there is a challenge in explaining the difference in performance between completely unconstrained networks (e.g., ParticleNet \cite{2020particlenet}) and highly expert-engineered networks (e.g., the learnable linear basis described in~\cite{2018efp}). Constructing equivariant (or partially equivariant) models can help  interpolate between these two edge cases, and understand which features are important in the task. This is clearly related to the qualities of interpretability and generalizability: presumably overparameterized networks are, to some degree, exploiting underlying symmetries of the system they are trained upon. However, if networks are not encouraged to generalize according to the available transformation properties then they will not explore or exploit these symmetries well. As shown in \Figref{boost_invariance}, jet tagging networks with Lorentz equivariance are robust to unseen boosting regimes, whereas non-equivariant designs may be sensitive to these transformations. The consequences of this sensitivity are also clearly shown in~\cite{gong2022efficient} (Figure 3 of that work), where the jet tagging performance of unconstrained networks diminishes for instance as input data is Lorentz-boosted. For models where Lorentz equivariance is enforced, generalization is almost perfect. It remains to be studied whether partially equivariant networks~\cite{Murnane:2022pmd} maintain this generalizability.

%-----------------------------
%-----------------------------

There is an open question of whether an unconstrained network can learn the symmetries of a physical system by being exposed to all possible transformations in the space during training. Certainly data augmentation with relevant symmetries can lead to improved performance~\cite{dillon2021symmetries}, although there remains to be done generalizability and interpretability studies of whether this augmentation is leading to a genuine learning of the underlying symmetries. It may be possible to use generative models to discover symmetries on various physics datasets \cite{symmetrygan}. An additional promising direction may be to recover symmetries by learning some latent representation of a physical system (for example as a Structured World Model~\cite{kipf2020contrastive}) and study the symmetry structure in this space~\cite{park2022learning}.
\enlargethispage{\baselineskip}

%=============================
\section{Reconsidering Evaluation Metrics  \label{sec:metrics}}
%=============================

% \editnote[Editors]{These are the major bullet points that we intend to address with this Section.    
%     \begin{itemize}
%         \item How to assess performance of these methods? 
%         \begin{itemize}
%             \item ``Ant factor'': size of model vs. strength of model \cite{Murnane:2022pmd}
%             \item GDP $!=$ happiness of citizens
%             \item How well the model generalizes
%             \item sample efficiency
%         \end{itemize}
%     \end{itemize}}
    
    The merits of a scientific method cannot be argued without first establishing common metrics by which to judge such methods. Traditionally, machine learning models are evaluated and compared based on simple performance metrics such as accuracy, error, true/false positive rates, receiver operating characteristic (ROC), and, to a lesser extent, computational efficiency. These metrics are indeed good at evaluating the raw performance of a \textit{statistical} model: they judge a machine learning model by ability to reproduce or extract certain features of the data. However, in many scientific applications, and especially in physics, a good description of a system needs to do more than merely reproduce observations. A good \textit{physical} model of a system should make accurate predictions while also satisfying several qualitative criteria. For example, it should minimize the number of free parameters needed to express the complexity of the system; these parameters should be chosen in a way that clarifies their physical meaning as much as possible; the values of the model's parameters can be learned from a dataset that is smaller than that needed for a generic model with many parameters; the model is generalizable and makes accurate predictions even for experiments that have never been conducted -- in other words, all of the qualities that equivariance is designed to enhance. These properties can act as a guide for extending a purely statistical model to a more comprehensive physical description of a system.

\begin{figure}
        \centering
        \includegraphics[width=0.7\linewidth]{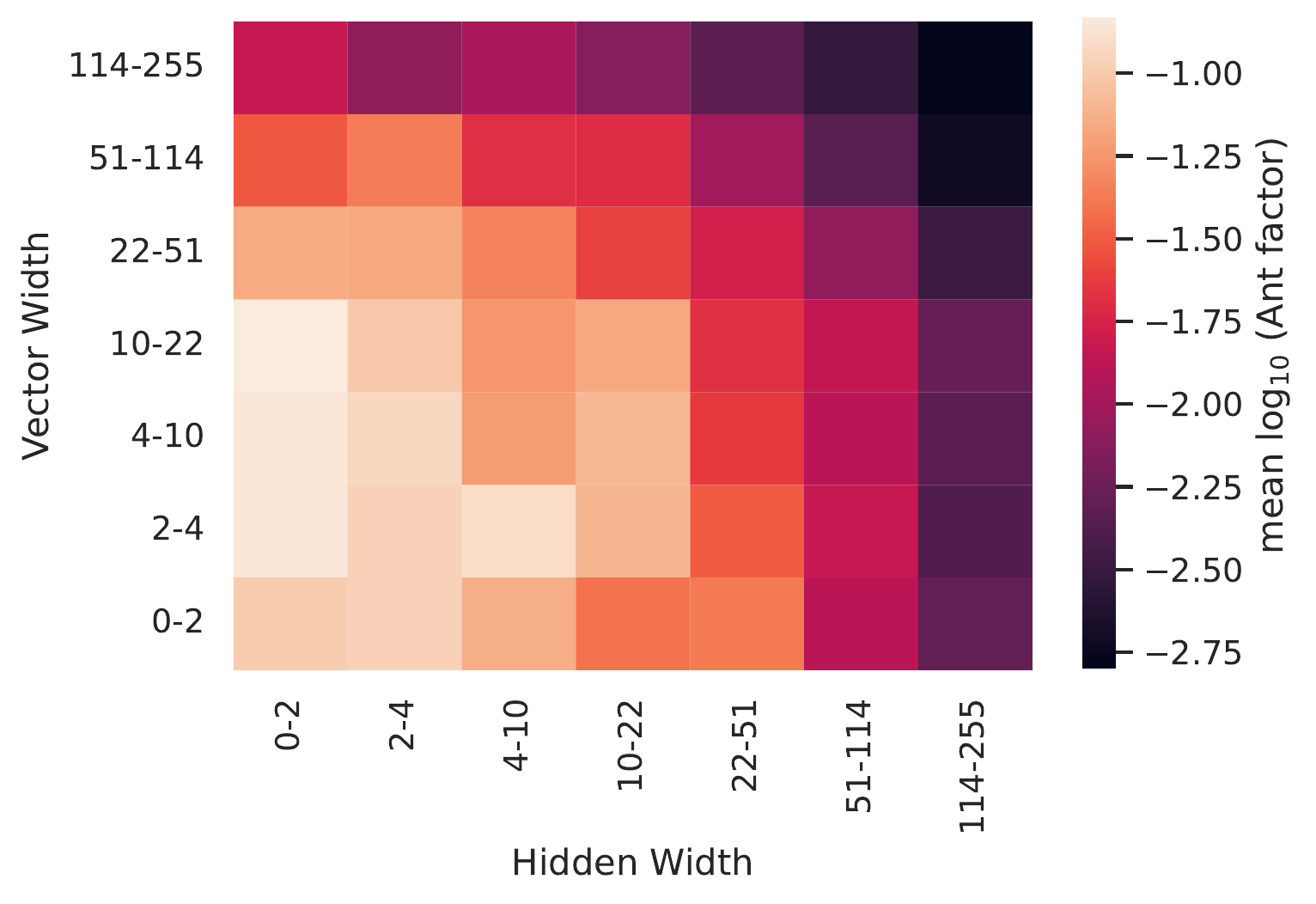}
        \caption{An example of quantifying model efficiency as a function of its symmetry-constrained properties. Unconstrained learning is along the x-axis, Lorentz-constrained learning is along the y-axis, and lighter colors show greater efficiency (as captured by the ``ant factor'', a power-to-weight ratio)
        % \editnote[Daniel]{To make this more cartoonish}
        }
        \label{fig:ant_factor}
    \end{figure}

    In light of this discussion, we propose that the community focus on developing and implementing a more holistic approach to evaluating and comparing machine learning models in scientific applications. Here we list only a few common ways of quantifying the above qualities of neural networks.
    %-----------------
    \begin{itemize}
        \item \textit{Quantifying model size}. Model size (and/or capacity) is fairly trivial to report, but is still often under-emphasized in the literature. The \textit{number of parameters} is the simplest metric of model size, but others metrics such as memory usage can also be considered. Lowering the number of parameters, assuming performance doesn't suffer, brings one closer to a viable physical model. The connection between model size and ease-of-training is non-trivial. While large unconstrained models need to learn symmetries themselves, they are also more likely to pick a "winning ticket", according to the Lottery Ticket Hypothesis~\cite{frankle2019lottery}. This suggests that it is useful to consider the effect of model size in symmetry-constrained networks separately in training and inference. 
        \item \textit{Quantifying inference efficiency}.
        A typical use-case for physics-informed ML is to miniaturize a model for inference on-the-edge in some experiment set-up. We suggest that some measure of power-to-weight ratio may be useful when surveying the wide variety of ML architectures and features applied to a physics task. In \cite{Murnane:2022pmd} one such measure is proposed: an ``ant factor'' that represents the performance of a network relative to its size. Here ``performance'' and ``size''  should be defined by the scientific field as is most relevant. An example of this ant factor (based on background rejection rate) is given in Figure  \ref{fig:ant_factor} as a function of the number of equivariant channels vs.~ non-equivariant channels. This sort of exploration could guide a community to an architecture most appropriate to its hardware budget or latency constraints.
        \item \textit{Quantifying sample efficiency}. Sample efficiency is commonly illustrated via \textit{learning curves}. Depending on the dataset, equivariant neural networks can achieve equal performance with only a fraction of the training data, and such comparisons should ideally be reported in publications.
        \item\textit{Quantifying variance/bias tradeoff.} Closely related to the above, since equivariant networks operate in a hypothesis space of reduced size, they are less likely to overfit the data, specifically they are prevented from overfitting the data in a way that would violate the underlying group symmetries.  
        \item \textit{Quantifying generalizability and interpretability}. These two properties are much more difficult to quantify, and it is a topic ongoing research in the machine learning community. \textit{Transfer learning}, the process of training a network in some region of phase space and gauging its performance in another, can be used to demonstrate generalization. Methods such as \textit{network dissection}~\cite{bau2017network}, originally suited for CNNs, can be extended to arbitrary equivariant architectures to provide metrics of interpretability. Notably, this may be easier to do in scientific applications where the tasks are typically less complex than in general purpose image recognition, for example.
    \end{itemize}
    %-----------------
    
    All this is not to say that pure model performance is unimportant, or that it needs to be sacrificed in favor of other properties. In fact, equivariant architectures already provide state-of-the-art predictions in scientific contexts, as seen, for example, in chemistry and high energy physics~\cite{AndeHyKon19,gong2022efficient}, and we believe that further research of equivariant architectures will only increase the performance gap between them and generic networks applied to physics problems. Nevertheless, the value of these models is \textit{not only} in their raw performance, and we hope that this paper will help convince other scientists in the community that a broader approach to network design that is less focused on short-term performance gains can facilitate bridge-building between machine learning and physics.

    %-----------------
    %-----------------

%=============================
\section{Soft Equivariance \label{sec:softequivariance}}
%=============================

Exact symmetries do not exhaust the possibilities for symmetry-informed models and problems. Many real-world problems deal only with approximate symmetries or combine symmetric and non-symmetric types of data. This issue comes up already in typical applications of CNNs for image recognition: the convolutional layers of CNNs provide translation-equivariant latent representations, but the network output emerges from a final, non-equivariant stage of the network (say, a dense layer). Moreover, images are always finite, and edge effects are a very well known issue for CNNs. The translational symmetry is realized only with respect to a subset of the full symmetry group. In machine learning, the relevant distinction is between soft and hard inductive biases. Up to this point, we have focused on the hard inductive bias introduced by a rigidly equivariant architecture. Here we briefly discuss the possibility of softer versions of that constraint. We can distinguish several types of soft or broken symmetry:
%-----------------
\begin{itemize}
    \item \textit{Infinitesimal symmetry.} Finite transformations may not always be a symmetry for a problem, but their infinitesimal versions might. For example, in image recognition and neuroscience, learning the infinitesimal \textit{generators} of transformations is common. Mathematically, such models work with representations of Lie algebras as opposed to groups.
    \item \textit{Partial symmetry.} By this we mean symmetry with respect to a subset of the full group of transformations (for example translations only up to a finite distance). In this case, approaches based on linear representation theory can still be effective because most of the structure of representations is determined just by the local neighborhood of the identity in the group. Including non-equivariant sub-networks also becomes reasonable.
    \item \textit{Approximate symmetry.} In many problems the symmetry transformations aren't expected to work in an exact manner at all -- this could be due to the presence of noise, interactions between different parts of the data, or even fundamental physical effects such as symmetry breaking. In these cases it is possible to relax the symmetry constraint gradually by allowing limited data mixing between different irreducible representations that wouldn't interact under exact symmetry.
\end{itemize}
%-----------------

One well-known example of a synthesis between equivariant and non-equivariant methods is AlphaFold~\cite{alphafold}, whose key component is an equivariant transformer~\cite{se3transformer}. Examples of models using partial equivariance for physics tasks include GNNs for jet tagging~\cite{Murnane:2022pmd} and residual networks for dynamical systems~\cite{finzi2021residual}.

%=============================
\section{Discussion \label{sec:discussion}}
%=============================

% \editnote[Editors]{Aims and brainstorming for the discussion in this section.    
% \begin{itemize}
%     \item Suggestions for what a research team should think about when considering introducing one of these approaches (``Should we use a BDT, a DNN, a CNN, a GNN, ...?''). What features of the system are you trying to extract? What is the task (classification, regression, etc.)? How much data do you have? 
%     \item Actionable guidance!
%     \item Make sure to consider potential drawbacks, as well as evaluating a model's value more holistically (is raw performance the only thing that matters? How about sample efficiency? etc.)
%     \begin{itemize}
%         \item What about particle physics, where we DO have nearly unlimited data? Why should we care about sample efficiency? (analogy to computer vision) 
%         \item Adding equivariance changes the scaling laws for a model's performance vs. number of samples: https://arxiv.org/abs/2101.03164v3
%     \end{itemize}
%     \item What are the specific use cases where equivariance could make a big difference? 
%     \begin{itemize}
%         \item End-to-end physics models (is this new physics or not?) 
%         \item Equivariant blocks within a classification/regression task: e.g. CP-violating decays
%         \item online inference/ML on the edge (need smaller models) 
%         \item uncovering new physics with semi-equivariant models? 
%     \end{itemize}
% \end{itemize}
% }

When developing a neural network approach for any task in physics, one should consider the specifics of the task when determining what kind of architecture to use: how the data is represented, which symmetry group(s) govern it, and what kind of information one is trying to extract. Below are a few practical points to keep in mind when considering the development and use of equivariant networks.

%-----------------
\begin{itemize}
    \item \textit{Development and training time:} At present, the main deterrent to a more widespread use of equivariant architectures is the time investment they require for model design, development, and training -- which includes the time possibly needed for the technical implementation of complicated mathematical symmetry group structures, as well as additional computing time: During the training process, the forward and backward propagation steps may be less optimized than those of simpler models and not benefit from optimized tensor operations as easily, thus significantly extending the required training time. However, equivariant models generally require significantly smaller training datasets than simpler models, potentially offsetting the training inefficiencies introduced by the more complex mathematical computations during training. Furthermore, the equivariant ``building blocks'' of such architectures are typically highly reusable, so it can be the case that the time-consuming early development for a given symmetry may not need to be repeated. Additionally, the development of equivariant models for physics tasks opens potential avenues for collaboration with the broader ML community as many existing architectures for enforcing Euclidean equivariance may be easily modified to accommodate more exotic symmetries.
    
    \item \textit{High performance computing:} For the full exploitation of equivariant networks -- and more generally machine learning for physics -- it will be important for physics-specific equivariant architectures to be optimized for deployment on exascale computing hardware. This will require some investment and recognition on the HPC side that not all applications of ML to physics will use ``out-of-the-box'' designs like CNNs, and so hardware benchmarking and acceptance tests on limited default architectures may not meet all of our community needs.
    
    \item \textit{Data dimensionality:} One should consider the dimensionality of the sample data used for training -- or more precisely, the dimensionality of the data versus that of its symmetry group. While the use of equivariant networks can improve sample efficiency, the added computational cost may outweigh these benefits, especially for low-dimensional symmetry groups. For example, in ~\cite{Boyda:2020hsi} it was found that imposing gauge symmetry significantly improved model efficiency, but imposing a residual hypercubic symmetry was expensive and did not improve performance. In addition, network training dynamics can be complicated when the training path is restricted to lie in symmetry-respecting space, and in some cases it may be more efficient to allow the network's optimisation trajectory to move through non-physical model space.
    
    \item \textit{Towards a general-use toolkit:} There has been significant effort spent in transferring physics-informed ML ideas from the computer vision community to the physics community (and indeed proving that these ideas are even applicable). As pointed out above, for a particular use-case, components of an equivariant architecture may be re-usable and transferable. However, these deep explorations of a particular symmetry should run in parallel with an attempt to generalize equivariance tools across symmetry groups, scientific fields, and core architectures (CNNs, GNNs, RNNs, transformers, etc.). Most open-source libraries for implementing equivariance are currently highly specialized in this regard. This is clearly not a trivial task, with several significant obstacles. Group calculations are usually performed symbolically (for example in Sympy~\cite{10.7717/peerj-cs.103}), then converted to convolutions or numerical formulae by hand. Libraries for this are specialized to particular families of symmetries and it would be a Herculean task to generalize these across multiple families of symmetry. A second obstacle that a general toolkit would face is that many of the highly optimized algorithms that enable unconstrained ML are non-trivial within an equivariant architecture. For example, nearest-neighbor algorithms are the backbone of many graph neural network approaches, and within Euclidean space there are dozens of libraries available. However, libraries for performing neighbor searches in curved or non-positive-definite space (e.g. spacetime) are essentially non-existent. Some headway may be made across these obstacles by use of approximations of symmetries, or low-order truncations of group operations (for example, by the relatively simple convolutions suggested in \cite{satorras2022en}).

\end{itemize}
%-----------------

%=============================
\section{Conclusions \label{sec:conclusions}}
%=============================

Symmetries underlie countless physical phenomena throughout the universe across all length and energy scales. This White Paper puts forth the arguments for incorporating knowledge of these symmetries and their equivariance properties into the algorithms and architectures that scientists use to investigate and describe numerous physical systems, with a particular emphasis on the potential for impacts in fundamental particle physics. Numerous advantages and potential limitations of such an approach are discussed, along with proposals for developing and implementing a more holistic approach to evaluating and comparing machine learning models in scientific applications. The investment of time and funding into the construction of equivariant architectures for physically-motivated symmetry groups is likely to have far-reaching benefits for applications across fundamental research in science and industry. The development of symmetry group equivariant machine learning architectures has substantial potential to aid in future physics discoveries, but only if we recognize its importance by dedicating the necessary resources to support it.

\bibliographystyle{JHEP}
\bibliography{references}

%%%%%%%%%%%%%%%%%%%%%%%%%%%%%%%%%%%%%%%%%

\end{document}